\documentclass[sigconf]{acmart}

\AtBeginDocument{%
  }

\copyrightyear{2024}
\acmYear{2024}
\setcopyright{rightsretained}
\acmConference[SIGIR '24]{Proceedings of the 47th International ACM SIGIR Conference on Research and Development in Information Retrieval}{July 14--18, 2024}{Washington, DC, USA}
\acmBooktitle{Proceedings of the 47th International ACM SIGIR Conference on Research and Development in Information Retrieval (SIGIR '24), July 14--18, 2024, Washington, DC, USA}
\acmDOI{10.1145/3626772.3657918}
\acmISBN{979-8-4007-0431-4/24/07}

\makeatletter
\gdef\@copyrightpermission{
\begin{minipage}{0.3\columnwidth}
\href{https://creativecommons.org/licenses/by/4.0/}{\includegraphics[width=0.90\textwidth]{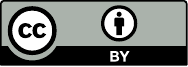}}
\end{minipage}\hfill
\begin{minipage}{0.7\columnwidth}
\href{https://creativecommons.org/licenses/by/4.0/}{This work is licensed under a Creative Commons
Attribution International 4.0 License.}
\end{minipage}
\vspace{5pt}
}
\makeatother

\usepackage{amsmath}
\usepackage{enumerate}
\usepackage{multirow}
\usepackage{natbib}
\usepackage{tabularx}
\usepackage{arydshln}
\usepackage[inline]{enumitem}
\usepackage{bbm}
\usepackage{mleftright}

\newcommand{\headernodot}[1]{\vspace{1mm}\noindent\textbf{#1}}
\newcommand{\header}[1]{\headernodot{#1.}}

\begin{document}

\title{Estimating the Hessian Matrix of Ranking Objectives for
Stochastic Learning to Rank with Gradient Boosted Trees}

\author{Jingwei Kang}
\orcid{https://orcid.org/0009-0003-9283-4060}
\affiliation{%
  \institution{University of Amsterdam}
  \city{Amsterdam}
  \country{The Netherlands}
}
\email{j.kang@uva.nl}

\author{Maarten de Rijke}
\orcid{https://orcid.org/0000-0002-1086-0202}
\affiliation{%
  \institution{University of Amsterdam}
  \city{Amsterdam}
  \country{The Netherlands}
}
\email{m.derijke@uva.nl}

\author{Harrie Oosterhuis}
\orcid{https://orcid.org/0000-0002-0458-9233}
\affiliation{%
  \institution{Radboud University}
  \city{Nijmegen}
  \country{The Netherlands}
}
\email{harrie.oosterhuis@ru.nl}

\begin{abstract}
    Stochastic learning to rank (LTR) is a recent branch in the LTR field that concerns the optimization of probabilistic ranking models.
    Their probabilistic behavior enables certain ranking qualities that are impossible with deterministic models.
    For example, they can increase the diversity of displayed documents, increase fairness of exposure over documents, and better balance exploitation and exploration through randomization.
    A core difficulty in LTR is gradient estimation, for this reason, existing stochastic LTR methods have been limited to differentiable ranking models (e.g., neural networks).
    This is in stark contrast with the general field of LTR where Gradient Boosted Decision Trees (GBDTs) have long been considered the state-of-the-art.

    In this work, we address this gap by introducing the first stochastic LTR method for GBDTs.
    Our main contribution is a novel estimator for the second-order derivatives, i.e., the Hessian matrix, which is a requirement for effective GBDTs.
    To efficiently compute both the first and second-order derivatives simultaneously, we incorporate our estimator into the existing PL-Rank framework, which was originally designed for first-order derivatives only.
    Our experimental results indicate that stochastic LTR without the Hessian has extremely poor performance, whilst the performance is competitive with the current state-of-the-art with our estimated Hessian.
    Thus, through the contribution of our novel Hessian estimation method, we have successfully introduced GBDTs to stochastic LTR.
\end{abstract}



\maketitle

\section{Introduction}
Learning to rank (LTR) is a core problem in the information retrieval (IR) field that concerns the optimization of ranking models~\cite{INR-016}. 
Generally, ranking models use a scoring function that independently assigns a score to each document to be ranked, and subsequently, ranks them by ordering them based on their assigned scores.
Traditionally, this ordering was done by deterministically sorting, i.e., the highest scored document is placed at the first rank, the second highest at the second rank, etc.
The aim of LTR techniques is to optimize a ranking metric based on the produced rankings, for instance, discounted cumulative gain (DCG), precision, or recall~\citep{jarvelin2002cumulated, harman2011information}.
The difficulty with this approach is that these metrics, and the sorting process underlying the ranking, are not differentiable~\citep{burges2010from, 10.1145/3404835.3462830, INR-016}.
Therefore, LTR has always relied on methods for gradient estimation or approximation~\citep{10.1145/3477495.3531842}.

The existing families of LTR methods can be divided into three categories according to their optimization objective: 
\begin{enumerate*}[label=(\roman*)]
\item heuristic approximations of ranking metrics~\citep{qin2010general, jagerman2022optimizing}; 
\item proven lower bounds on ranking metrics~\citep{10.1145/1341531.1341544, 10.1145/1277741.1277790, 10.1145/1390334.1390355, 10.1145/1401890.1401906}; or 
\item the expected values of ranking metrics under stochastic ranking models~\citep{10.1145/3340531.3411962, 10.1145/3336191.3371844, NEURIPS2019_9e82757e, 10.1145/3404835.3462830}. 
\end{enumerate*}
The latter often choose to optimize Plackett-Luce (PL) ranking models as their stochastic rankers~\citep{10.1145/3404835.3462830, 10.1145/3477495.3531842, af5079a1-8ca5-3727-a405-0a82390327b7, luce_individual_1959, xia2008listwise, cao2007learning, ustimenko2020stochasticrank}.
The advantage of optimizing the expected values of ranking metrics is that they are fully differentiable, unlike their deterministic counterparts.
However, this stochastic LTR approach comes with two important challenges:
\begin{enumerate*}[label=(\roman*)]
    \item If the goal is to produce a deterministic model then it may not be perfectly aligned with the stochastic LTR objective;
    and
    \item exact computation of an expected value requires iterating over all possible rankings, i.e., every permutation of documents, which is infeasible in practice~\citep{10.1145/3404835.3462830}.
\end{enumerate*}

\begin{figure*}[tp]
\centering
\begin{equation*}
\begin{split}
\qquad
\qquad
\frac{\partial^2}{\partial m(d)^2}R(\pi) 
= \mathbb{E}_y \Biggr[  &\Biggr( 1 - \sum_{x=1}^{\text{rank}(d,y)}\pi(d\mid y_{1:x-1}) \Biggr) \Biggr(\sum_{k=\text{rank}(d,y)+1}^{K} \theta_k\rho_{y_k}\Biggr) \\
& + \sum_{k=1}^{\text{rank}(d,y)}\pi(d\mid y_{1:k-1}) \Biggr(\theta_k\rho_{d}-\sum_{x=k}^{K}  \theta_x\rho_{y_x} \Biggr) \Biggr(\mathbbm{1}[d \in y ] + 1 - \pi(d\mid y_{1:k-1}) - \sum_{x=1}^{\text{rank}(d,y)}\pi(d\mid y_{1:x-1}) \Biggr)\Biggr]
\end{split}
\end{equation*}
\vspace{-\baselineskip}
\caption{The complete second-order derivative of the reward function $R(\pi)$ w.r.t.\ the scoring function $m(d)$.}
\vspace{-\baselineskip}
\label{fig: Hessian}
\end{figure*}

Several methods have been designed for the latter issue.
Early listwise LTR work optimizes PL ranking models, and avoids the permutation problem by only maximizing the probability of a single optimal ranking~\citep{xia2008listwise, cao2007learning}.
More recent stochastic LTR work uses sampling to approximate the gradient instead.
\citet{10.1145/3336191.3371844} proposed to sample rankings from PL models using the computationally efficient Gumbel-Softmax trick~\cite{gumbel1948statistical}.
\citet{NEURIPS2019_9e82757e} applied a policy-gradient method to approximate the gradient based on sampled rankings.
Subsequently, \citet{10.1145/3404835.3462830} proposed the PL-Rank method that makes use of several mathematical properties of ranking metrics and PL ranking models to perform gradient estimation with high computational efficiency~\citep{10.1145/3477495.3531842}.
Stochastic LTR with PL-Rank converges significantly quicker and at higher levels of performance than standard policy-gradient approaches.

Besides these different high-level approaches, the current state of stochastic LTR has another interesting difference with deterministic LTR.
To the best of our knowledge, all stochastic LTR works optimize neural networks (NNs)~\citep{10.1145/3404835.3462830, 10.1145/3477495.3531842, af5079a1-8ca5-3727-a405-0a82390327b7, luce_individual_1959, Buchholz2022},\footnote{Technically, \citep{ustimenko2020stochasticrank} is an exception to this observation.}
whilst in deterministic LTR, Gradient Boosted Decision Trees (GBDTs) have historically had the most prominent role~\citep{burges2010from, 50030, dato:sigir2022-istella, 10.1145/2939672.2939785}.
Accordingly, it is surprising that despite their enormous importance to the LTR field, GBDTs have been largely ignored for stochastic LTR.

This work aims to bridge this gap;
We first point out that existing stochastic LTR work has been focused on estimating the first-order derivatives of ranking objectives.
But GBDTs also require the second-order derivatives for their optimization, as a result, the existing methods are not applicable to GBDTs~\citep{10.1145/3404835.3462830, 10.1145/3477495.3531842, af5079a1-8ca5-3727-a405-0a82390327b7, luce_individual_1959}.
We address this issue by proposing a novel estimator for these second-order derivatives, i.e., Hessian matrices, in a computationally efficient manner.
Our method integrates very well with PL-Rank~\citep{10.1145/3404835.3462830, 10.1145/3477495.3531842}, such that both first and second-order derivatives are computed efficiently.
Our experimental results show that stochastic LTR via GBDTs without the Hessian leads to detrimental performance.
Conversely, with our estimated Hessian, GBDTs are able to outperform NNs on several LTR benchmarks by considerable margins.
Furthermore, we observe that stochastic LTR with NNs does not have stable convergence, and thus requires early-stopping, but we do not observe such behavior for GBDTs with our estimated Hessians.
Thus, by contributing the first Hessian estimation method for stochastic LTR, we have closed an important gap with deterministic LTR, that enables substantial performance and stability improvements.

\section{Background: Stochastic LTR}
\header{PL ranking models}
Plackett-Luce ranking models provide a probability distribution over fixed-length permutations~\citep{af5079a1-8ca5-3727-a405-0a82390327b7, luce_individual_1959}. 
Accordingly, they have often been deployed as models of stochastic ranking models~\citep{10.1145/3404835.3462830, 10.1145/3477495.3531842, af5079a1-8ca5-3727-a405-0a82390327b7, luce_individual_1959, xia2008listwise, cao2007learning, ustimenko2020stochasticrank}. 
We use $y$ to denote a ranking $y=[y_1,y_2,\ldots,y_K]$ with a cutoff of $K$, and $y_{1:k}=[y_1,y_2,\ldots,y_k]$ the partial ranking up to rank $k$.
Let $\pi$ indicate a Plackett-Luce ranking model, with $m(d)$ representing the log score of document $d$, then the probability that $d$ is chosen to be the $k$-th item in ranking $y$ from the set of items $D$ is:
\begin{equation}
\pi(d\mid y_{1:k-1},D) = \frac{e^{m(d)}\mathbbm{1}[d \notin y_{1:k-1}]}{\sum_{d' \in D \setminus y_{1:k-1}} e^{m(d')}}.
\end{equation}
The probability of the overall ranking $y$ is the product of the placement probabilities of each individual document:
\begin{equation}
\pi(y) = \prod_{k=1}^K\pi(y_k\mid y_{1:k-1},D)
.
\end{equation}
\header{Ranking metrics}
Our objective is to optimize a ranking metric that fits the form of a $DCG@K$ metric~\citep{jarvelin2002cumulated}.
This means that each document $d$ has a relevance $\rho_d$ and each rank has a weight $\theta_k$, e.g., for standard DCG: $\theta_{k} =  \frac{\mathbbm{1}[ k\leq K ]}{\log_2(k+1)}$; the expected value of this metric under $\pi$ is then:
\begin{equation}
R(\pi) =\sum_{y \in \pi} \pi(y) \sum_{k=1}^{K}\theta_{k}\rho_{y_k} = \mathbb{E}_y \Biggl[ \sum_{k=1}^{K}\theta_{k}\rho_{y_k} \Biggr].
\label{eq:reward}
\end{equation}

\header{PL-Rank}
\citet{10.1145/3404835.3462830} found that the gradient of $R(\pi)$ w.r.t. the general scoring function $m(d)$ can be expressed as:
\begin{equation}
\begin{split}
\frac{ \partial }{ \partial m(d)}R(\pi) ={}& \mathbb{E}_y \Biggl[ \Biggl( \sum_{k=\operatorname{rank}(d,y)+1}^{K} \theta_k\rho_{y_k} \Biggr) \\
&{} +  \sum_{k=1}^{\operatorname{rank}(d,y)} \pi(d \mid y_{1:k-1}) \Biggl( \theta_k\rho_d - \sum_{x=k}^{K} \theta_x\rho_{y_x} \Biggr) \Biggr].
\end{split}
\label{original plrank}
\end{equation}
Subsequently, \citet{10.1145/3477495.3531842} introduced the PL-Rank algorithm (specifically PL-Rank-3) to compute this gradient from a set of $N$ sampled rankings, with high computational efficiency and sample efficiency. 
For each sampled ranking $y$, PL-Rank starts by pre-computing the following values (in linear time, $\mathcal{O}(K)$):
\begin{equation}
\begin{split}
{PR}_{y,i} &= \sum_{k=i}^{K} \theta_k \rho_{y_k}, \quad {PR}_{y,d}={PR}_{y,\operatorname{rank}(d,y)+1},\\
{RI}_{y,i} &= \hspace{-0.2cm} \sum_{k=1}^{\min(i,K)}\hspace{-0.2cm} \frac{PR_{y,k}}{\sum_{d' \in D \setminus y_{1:k-1}} e^{m(d')}}, \quad {RI}_{y,d}={RI}_{y,\operatorname{rank}(d,y)},\\
{DR}_{y,i} &= \hspace{-0.2cm} \sum_{k=1}^{\min(i,K)}\hspace{-0.2cm} \frac{\theta_k}{\sum_{d' \in D \setminus y_{1:k-1}} e^{m(d')}}, \quad {DR}_{y,d}={DR}_{y,\operatorname{rank}(d,y)}.
\label{simplification}
\end{split}
\end{equation}
With these values pre-computed, gradient computation for a single ranking can also be done in linear time since:
\begin{align}
\frac{ \partial }{ \partial m(d)}R(\pi)
&= \mathbb{E}_{y} \Biggr[ PR_{y,d} + e^{m(d)}\Biggl( \rho_d DR_{y,d} -RI_{y,d}\Biggl) \Biggr]
\label{plrank3}\\
&\approx \frac{1}{N}\sum_{i=1}^{N} \biggr[ PR_{y^{(i)},d} + e^{m(d)}\biggl( \rho_d DR_{y^{(i)},d} -RI_{y^{(i)},d}\biggl) \biggr],
\nonumber
\end{align}
where the second line shows the gradient estimator of PL-Rank.
Because this has to be done for each of the $N$ sampled rankings, and each ranking is created through (top-$K$) sorting, the complexity of the entire algorithm is $\mathcal{O}(N \cdot (K + D))$.
Thereby, PL-Rank efficiently approximates the first-order derivatives from sampled rankings.

\section{Method: Approximating the Hessian}
The goal of this work is to make GBDTs work effectively for stochastic LTR.
As discussed earlier, GBDTs require both first- and second-order derivatives, and to the best of our knowledge, the stochastic LTR field is currently missing an algorithm for efficiently estimating the second-order derivatives.
Accordingly, as PL-Rank has done for the first-order derivatives, we contribute a novel algorithm to computationally-efficiently estimate the second-order derivatives of ranking metrics w.r.t.\ PL ranking models.

Our first step is to discover a formulation of the second-order derivatives. Luckily, the first-order derivatives in Eq.~\ref{original plrank} provides a useful starting point for the derivation.
Due to space limitations, we omit the intermediate steps of the derivation and only display the result in Figure~\ref{fig: Hessian}.
However, we believe that with the same starting point (Eq.~\ref{original plrank}) and knowledge of the result (Figure~\ref{fig: Hessian}), one can reproduce a derivation with moderate effort.

The full formula in Figure~\ref{fig: Hessian} does not immediately reveal an efficient manner of computing it.
Inspired by the PL-Rank approach, we aim to reformulate the second-order derivatives in terms that can be computed with minimal computational complexity.
Conveniently, we can reuse the pre-computed terms for PL-Rank in Eq.~\ref{simplification} for the second-order derivatives.
In addition, we introduce three more terms that can be pre-computed; the first is the cumulative sum of all reciprocal scoring denominators:
\begin{equation}
{DN}_{y,i} = \hspace{-0.3cm} \sum_{k=1}^{\min(i,K)}\hspace{-0.35cm} \frac{1}{\sum_{d' \in D \setminus y_{1:k-1}} e^{m(d')}}, \quad {DN}_{y,d}={DN}_{y,\operatorname{rank}(d,y)}.
\end{equation}
The other two terms are variations on $RI$ and $DR$ where the denominators have been squared:
\begin{equation}
\begin{split}
{RS}_{y,i} &= \hspace{-0.3cm} \sum_{k=1}^{\min(i,K)}\hspace{-0.35cm} \frac{PR_{y,k}}{(\sum_{d' \in D \setminus y_{1:k-1}} e^{m(d')})^2}, \quad {RS}_{y,d}\hspace{-0.05cm}=\hspace{-0.05cm}{RS}_{y,\text{rank}(d,y)},\\
{DS}_{y,i} &= \hspace{-0.3cm} \sum_{k=1}^{\min(i,K)}\hspace{-0.35cm} \frac{\theta_k}{(\sum_{d' \in D \setminus y_{1:k-1}} e^{m(d')})^2}, \quad {DS}_{y,d}\hspace{-0.05cm}=\hspace{-0.05cm}{DS}_{y,\text{rank}(d,y)}.
\end{split}
\end{equation}
Importantly, these three terms can be computed with linear complexity, $\mathcal{O}(K)$, for a single ranking.
Furthermore, we introduce two functions, one that has to be multiplied with $e^{m(d)}$ and the other with $e^{m(d)^2}$:
\begin{equation}
\begin{split}
X_1(y,d) &=  \Big(1 + \mathbbm{1}[d \in y] \Big) \Big( \rho_d DR_{y,d} -RI_{y,d} \Big)
- DN_{y,d} PR_{y,d},
\\
X_2(y,d) &= 
\Big( RS_{y,d} - \rho_d DS_{y,d} \Big)
- DN_{y,d} \Big( \rho_d DR_{y,d} -RI_{y,d} \Big),
\end{split}
\end{equation}
where $\mathbbm{1}[d \in y]$ indicates whether $d$ is ranked in the top-$K$ in $y$.
Since the previous terms can be pre-computed and stored in $\mathcal{O}(K)$, subsequently, the computation of $X_1$ and $X_2$ is constant per document, therefore, computing them for \emph{all} documents gives a complexity of $\mathcal{O}(K + D)$.
Finally, the hessian can be expressed in these terms and estimated from $N$ sampled rankings accordingly:
\begin{align}
&\frac{ \partial^2}{ \partial m^2(d)}R(\pi) = \mathbb{E}_{y} \Biggr[ PR_{y,d} + e^{m(d)}X_1(y,d) + e^{m(d)^2}X_2(y,d) \Biggr],
\nonumber \\
&\quad \approx \frac{1}{N} \sum_{i=1}^N \Biggr[ PR_{y^{(i)}\!\!,d} + e^{m(d)}X_1(y^{(i)}\!\!,d) + e^{m(d)^2}X_2(y^{(i)}\!\!,d) \Biggr].
\label{eq:ourestimator}
\end{align}
Again, we see that after pre-computing the terms, the operation per document and ranking is constant.
As a result, computing the second-order derivative for every document, i.e., the Hessian, with our algorithm has the same complexity as the underlying sorting procedure used to sample rankings: $\mathcal{O}(N \cdot (K + D))$.

Our Hessian estimator integrates well with the existing PL-Rank algorithm since it reuses all of its pre-computed terms.
Similarly, the rankings sampled for estimating the first-order derivatives can also be reused for estimating the Hessian.
Therefore, we argue that it can be seen as a natural extension of the PL-Rank that makes it relevant to GBDTs.
Accordingly, just as PL-Rank, our Hessian estimator can also be used for optimizing the second-order derivatives of other types of ranking metrics, i.e., exposure-fairness ~\citep{10.1145/3404835.3462830}.
The remainder of this paper evaluates whether our Hessian estimator improves GBDTs optimization for relevance ranking metrics.

\section{Experimental Setup}

Our experiments answer the following two research questions:
\begin{enumerate}[align=left, label={\bf RQ\arabic*},leftmargin=*]
\item Do our estimated Hessians provide an increase in NDCG performance for stochastic LTR with GBDTs?
\label{rq:hessian}
\item Do GBDTs with our estimated Hessians reach higher levels of NDCG performance than NNs for stochastic LTR?
\label{rq:performance}
\end{enumerate}
Our experiments compare the follow three models:
\begin{enumerate*}[label=(\roman*)]
\item NNs optimized with PL-Rank~\citep{10.1145/3404835.3462830, 10.1145/3477495.3531842};
\item GBDTs optimized with gradients from PL-Rank and Hessians from our novel estimator (Eq.~\ref{eq:ourestimator});
and \item GBDTs optimized gradients from PL-Rank but without a Hessian (set to a value of $1$ for all documents).
\end{enumerate*}
Finally, to investigate the gap between stochastic and deterministic LTR, we also report the performance of the XGBoost built-in LambdaMART.

Our experiments use three publicly-available LTR datasets: Yahoo! Webscope-Set1~\cite{pmlr-v14-chapelle11a}, MSLR-Web30K~\cite{DBLP:journals/corr/QinL13}, and Istella~\cite{10.1145/2987380}.
We consider two ranking lengths $K=5$ and $K=10$ and optimize DCG$@K$; our evaluation metric is normalized DCG$@K$ (NDCG$@K$); each reported result is the mean of five independent runs.
The NNs were optimized with a PyTorch~\cite{NEURIPS2019_bdbca288} implementation of PL-Rank (Eq.~\ref{plrank3}), the GBDTs were implemented with XGBoost~\cite{10.1145/2939672.2939785}.
For hyperparameter tuning,  we used Optuna~\cite{10.1145/3292500.3330701}; for a fair comparison, each method was given exactly 12 hours of hyperparameter tuning on the validation set, per setting. Our experimental implementation is publicly available at \url{https://github.com/jkang98/2024-SIGIR-XGBoost-PL-Rank}.

\begin{table}
\caption{
NDCG$@K$ reached by three stochastic LTR methods and LambdaMART.
Results are means over five runs, with standard deviations shown in brackets.
Best performance by a stochastic method shown in bold.
}
\label{tab:mean_std}
\vspace{-\baselineskip}
\begin{tabularx}{\columnwidth}{c X l l}
\hline
\hspace{0.4cm}
& \multicolumn{1}{ c }{Method}
& \multicolumn{1}{ c }{$K=5$}
& \multicolumn{1}{ c }{$K=10$}
\\
\hline
\multirow{4}{*}{\rotatebox[origin=c]{90}{Yahoo!}}
& Neural Network
& 0.7566 {\tiny (0.0006)}
& 0.7867 {\tiny (0.0003)}
\\ 
& GBDT w/. Hessian
& \bf 0.7614 {\tiny (0.0012)}
& \bf 0.7905 {\tiny (0.0009)}
\\ 
& GBDT w/o. Hessian
& 0.7233 {\tiny (0.0004)}
& 0.7620 {\tiny (0.0012)}
\\ \cdashline{2-4}
& LambdaMART
& 0.7796
& 0.8035
\\ \hline
\multirow{4}{*}{\rotatebox[origin=c]{90}{MSLR}}
& Neural Network
& 0.4810 {\tiny (0.0023)}
& 0.4895 {\tiny (0.0017)}
\\ 
& GBDT w/. Hessian
& \bf 0.4844 {\tiny (0.0072)}
& \bf 0.4941 {\tiny (0.0054)}
\\ 
& GBDT w/o. Hessian
& 0.4313 {\tiny (0.0032)}
& 0.4429 {\tiny (0.0033)}
\\ \cdashline{2-4}
& LambdaMART
& 0.4889 {\tiny (0.0014)}
& 0.4955 {\tiny (0.0033)}
\\ \hline
\multirow{4}{*}{\rotatebox[origin=c]{90}{Istella}}
& Neural Network
& \bf 0.6117 {\tiny (0.0016)}
& \bf 0.6533 {\tiny (0.0011)}
\\ 
& GBDT w/. Hessian 
& 0.5942 {\tiny (0.0023)}
& 0.6376 {\tiny (0.0019)}
\\ 
& GBDT w/o. Hessian
& 0.5734 {\tiny (0.0057)}
& 0.5777 {\tiny (0.0034)}
\\ \cdashline{2-4}
& LambdaMART
& 0.6554
& 0.7018
\\ \hline
\end{tabularx}
\vspace{-\baselineskip}
\end{table}

{\renewcommand{\arraystretch}{0.5}
\begin{figure*}[tb]
\centering
\begin{tabular}{r r r c}
 \multicolumn{1}{c}{  \hspace{0.75cm} Yahoo! Webscope-Set1}
&
 \multicolumn{1}{c}{  \hspace{0.3cm} MSLR-Web30k}
&
 \multicolumn{1}{c}{  \hspace{0.2cm} Istella}
 &
 \multirow{4}{*}{\vspace{-4.5cm}\includegraphics[scale=0.3]{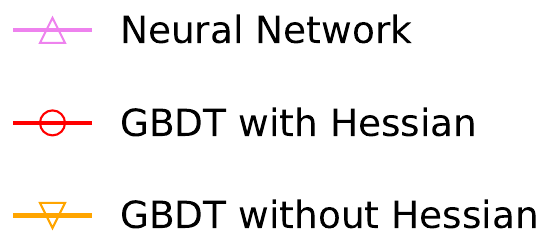}}
\\
\rotatebox[origin=lt]{90}{\hspace{0.05cm} \small NDCG@5 ($K=5$)} 
\includegraphics[scale=0.3]{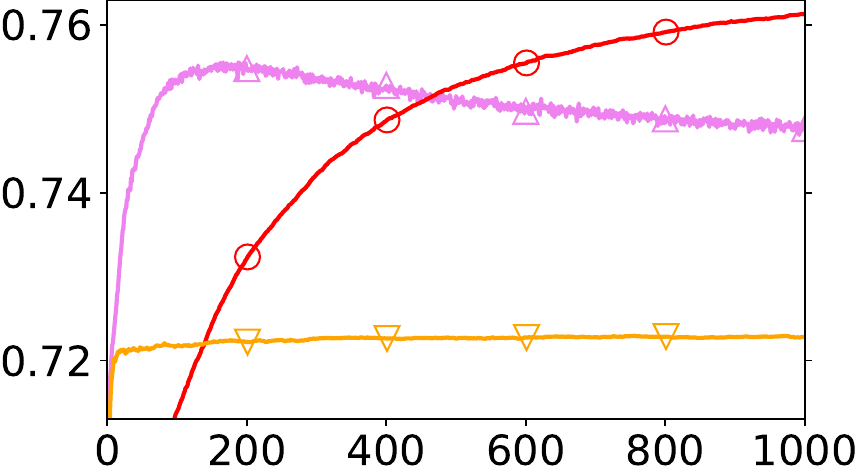}&
\includegraphics[scale=0.3]{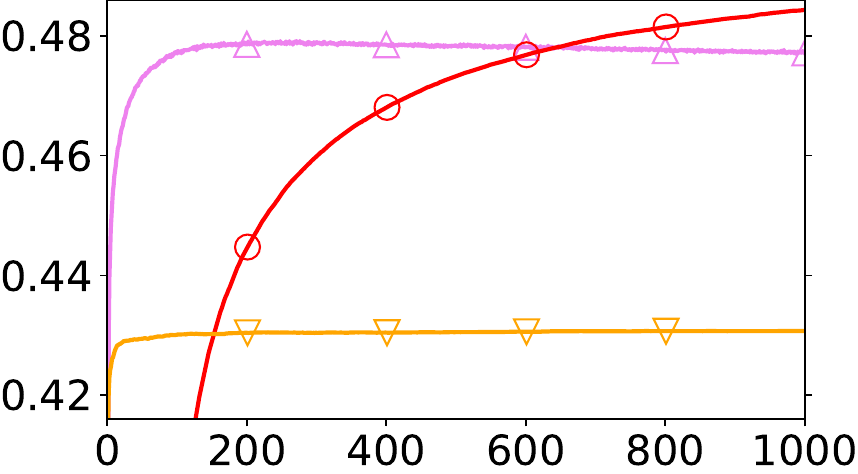} &
\includegraphics[scale=0.3]{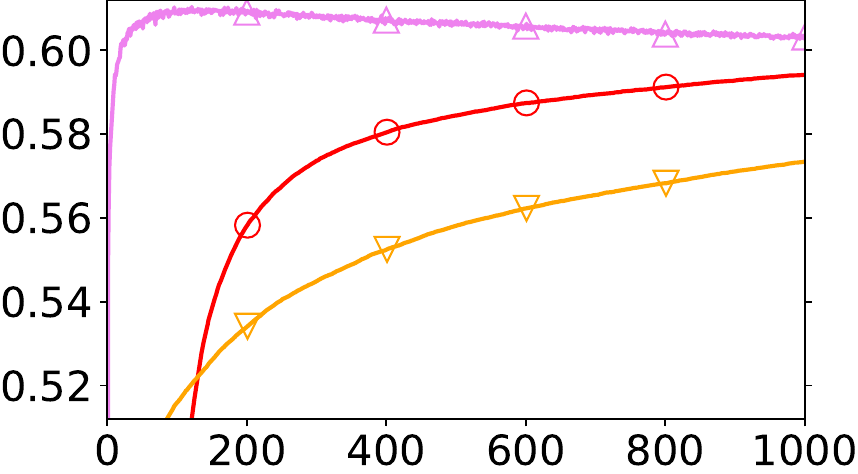} &
\\
\rotatebox[origin=lt]{90}{\hspace{-0.1cm} \small NDCG@10 ($K=10$)} 
\includegraphics[scale=0.3]{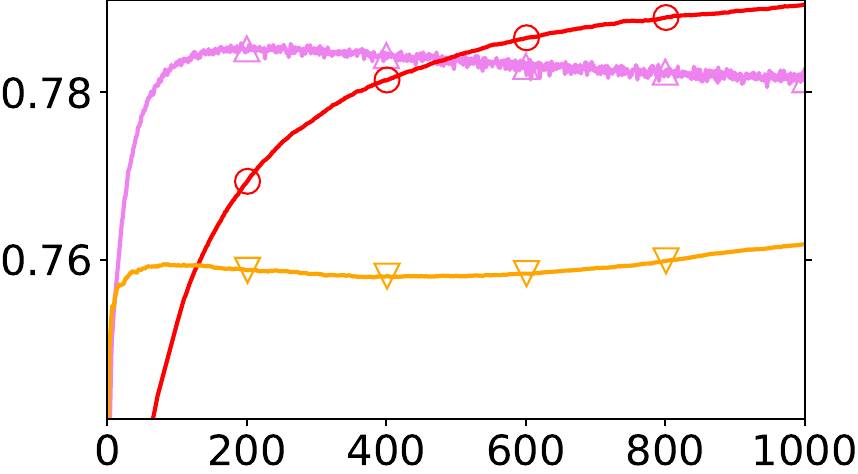}&
\includegraphics[scale=0.3]{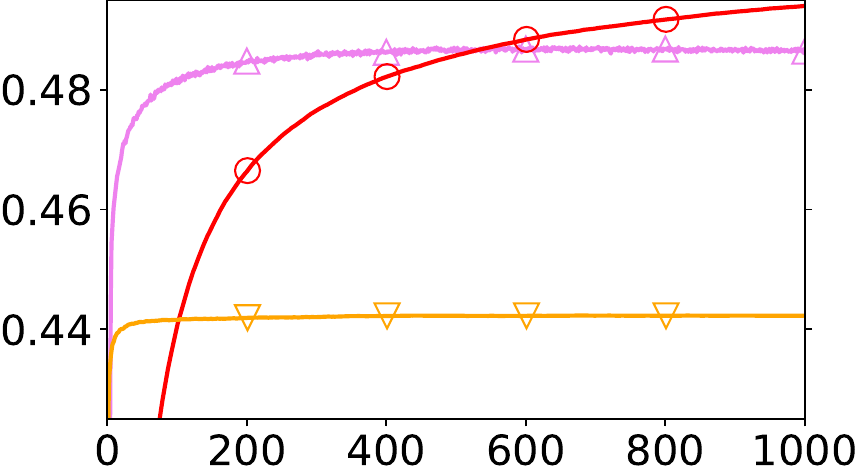} &
\includegraphics[scale=0.3]{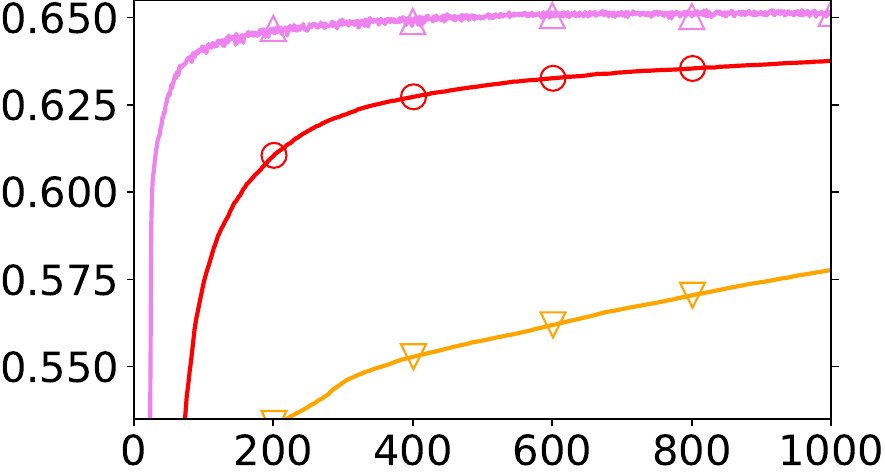} &
\\
 \multicolumn{1}{c}{ \hspace{2em} Epochs Trained}
& \multicolumn{1}{c}{ \hspace{1em} Epochs Trained}
& \multicolumn{1}{c}{ \hspace{1em} Epochs Trained} &
\end{tabular}
\vspace{-\baselineskip}
\caption{Learning curves of stochastic LTR methods in terms of NDCG$@K$ over a thousand epochs.
}
\vspace{-\baselineskip}
\label{fig:perf}
\end{figure*}
}

{\renewcommand{\arraystretch}{0.1}
\begin{figure}[tb]
\centering
\begin{tabular}{c}
\small \hspace{2em} Yahoo! Webscope-Set1
\\
\rotatebox[origin=lt]{90}{\hspace{0.1cm} \small NDCG@5 ($K=5$)} 
\includegraphics[scale=0.3
]{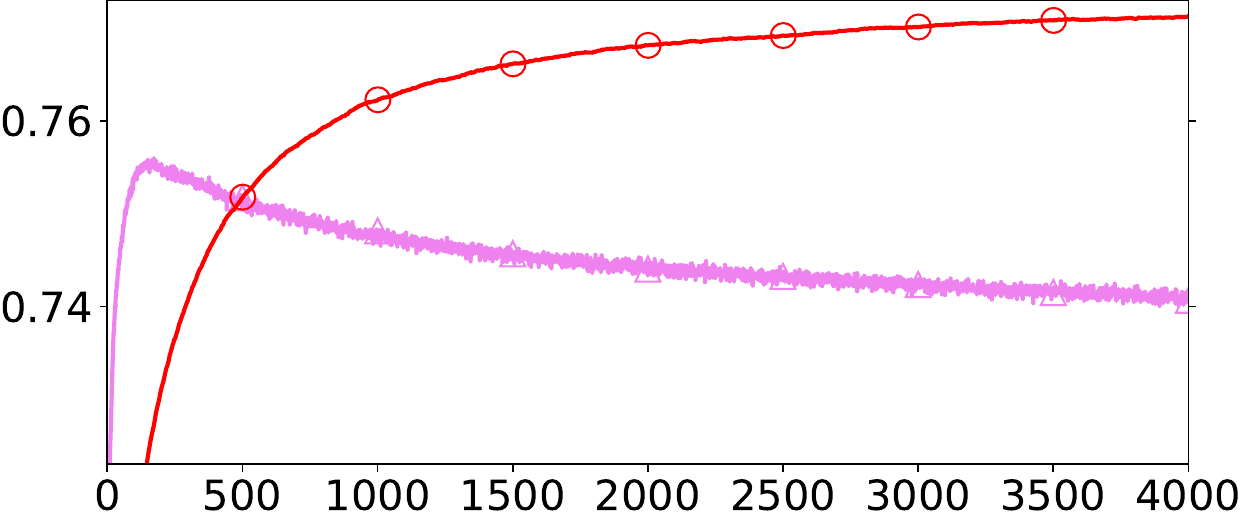}
\\
\small \hspace{0.5em} Epochs Trained
\\
\includegraphics[scale=0.38]{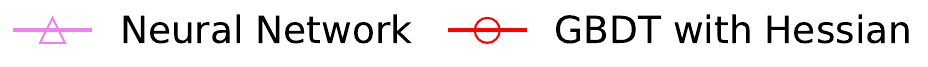}
\\
\end{tabular}
\vspace{-\baselineskip}
\caption{
Convergence analysis on the Yahoo!\ dataset.
}
\label{fig:converge}
\vspace{-\baselineskip}
\end{figure}
}

\section{Results}

\subsection{Importance of Hessian estimation}
We first consider \ref{rq:hessian}: \emph{whether our estimated Hessians provide an increase in NDCG performance for stochastic LTR with GBDTs.}
Table~\ref{tab:mean_std} displays the NDCG$@5/10$ reached by GBDTs with and without an estimated Hessian.
We see that, across all datasets and both ranking lengths, GBDTs without a Hessian reach a considerably lower NDCG than with our estimated Hessian.
The differences in performance range from $0.02$ to $0.06$, where the lower value of $0.02$ is already a very large difference for the NDCG metric.
Additionally, Figure~\ref{fig:perf} displays the learning curves of both methods.
Very clearly, GBDTs without Hessian consistently converge at suboptimal performance, with the exception on Istella dataset, this convergence takes place very quickly.
In contrast, with our estimated Hessians, GBDTs reach much higher performance and continue improving over many more epochs.

We answer \ref{rq:hessian} accordingly: Our estimated Hessian enables stochastic GBDT models to reach considerably higher NDCG than without a Hessian.
This demonstrates both the importance of Hessians for stochastic LTR and the effectiveness of our novel estimator.

\subsection{Comparison of GBDT and NN models}
Next, we address \ref{rq:performance}: \emph{whether GBDTs with our estimated Hessian reach higher levels of
NDCG performance than NNs}.
Table~\ref{tab:mean_std} reveals that this is clearly the case on Yahoo!\ and MSLR, but on Istella NNs have higher performance.
In Figure~\ref{fig:perf}, we see the learning curves of both methods.
Surprisingly, NNs appear to learn in fewer epochs but their performance degrades over time.
In contrast, GBDTs require more epochs but do not display such unstable convergence.
To further analyze convergence stability, we performed an additional experiment over 4000 epochs, the results of which can be seen in Figure~\ref{fig:converge}.
There, NN performance continues to degrade while the GBDTs continue to learn.
Thereby, our results suggest that GBDTs provide far stabler optimization that can reach higher performance on the Yahoo!\ and MSLR datasets.
It is unclear why similar results are not achieved on the Istella dataset; it is possible that more epochs are required for GBDTs to reach NN levels of performance there, but we could not confirm this.

Therefore, we answer \ref{rq:performance} as follows:
our estimated Hessian enables GBDTs to reach performance comparable to NNs, on the Yahoo!\ and MSLR datasets this lead to substantial improvements in NDCG, whilst on Istella, NN performance was not reached.
Additionally, we found that NNs do not converge stably and depend on early stopping for good performance.
Conversely, GBDTs have very stable convergence and no performance decreases were observed.

Finally, to evaluate the gap between stochastic and deterministic LTR, we also compare with the built-in LambdaMART implementation of XGBoost in Table~\ref{tab:mean_std}.
Unsurprisingly, built-in LambdaMART still outperforms our stochastic methods; this implementation benefits from almost a decade of implementation optimization.
Furthermore, while the GBDTs are optimizing distributions over rankings, the deterministic LambdaMART has a simpler task by focusing on individual rankings.
Nonetheless, on the Yahoo!\ and MSLR datasets, GBDTs with estimated Hessian get much closer to LambdaMART performance than the NNs.
In particular, the differences on MSLR between our GBDTs and LambdaMART are marginal.
Thereby, we believe our GBDTs provide an important step in bridging the gap between stochastic LTR and state-of-the-art deterministic LTR.

\section{Conclusion}
In this work, we introduced the first stochastic LTR method for effectively optimizing GBDTs.
We proposed a novel estimator for the second-order derivatives (i.e., the Hessian) of stochastic ranking objectives w.r.t.\ PL ranking models, and showed how it can be computed with minimal computational complexity.
Our experimental results reveal that GBDTs have extremely poor performance when optimized without Hessian.
Conversely, with our estimated Hessian, GBDTs are able to provide substantial performance gains over NNs.
Furthermore, unlike NNs, GBDTs have very stable convergence that enable the performance gains.
Our work brings stochastic LTR significantly closer to state-of-the-art deterministic LTR.

\subsubsection*{\bf Acknowledgements}
This work is partially supported by the Dutch Research Council (NWO), grant number VI.Veni.222.269 and project numbers 024.004.022, NWA.1389.20.183, KICH3.LTP.20.006.

\balance
\bibliographystyle{ACM-Reference-Format}
\bibliography{references.bib}


\begin{thebibliography}{30}


\ifx \showCODEN    \undefined \def \showCODEN     #1{\unskip}     \fi
\ifx \showDOI      \undefined \def \showDOI       #1{#1}\fi
\ifx \showISBNx    \undefined \def \showISBNx     #1{\unskip}     \fi
\ifx \showISBNxiii \undefined \def \showISBNxiii  #1{\unskip}     \fi
\ifx \showISSN     \undefined \def \showISSN      #1{\unskip}     \fi
\ifx \showLCCN     \undefined \def \showLCCN      #1{\unskip}     \fi
\ifx \shownote     \undefined \def \shownote      #1{#1}          \fi
\ifx \showarticletitle \undefined \def \showarticletitle #1{#1}   \fi
\ifx \showURL      \undefined \def \showURL       {\relax}        \fi
\providecommand\bibfield[2]{#2}
\providecommand\bibinfo[2]{#2}
\providecommand\natexlab[1]{#1}
\providecommand\showeprint[2][]{arXiv:#2}

\bibitem[Akiba et~al\mbox{.}(2019)]%
        {10.1145/3292500.3330701}
\bibfield{author}{\bibinfo{person}{Takuya Akiba}, \bibinfo{person}{Shotaro Sano}, \bibinfo{person}{Toshihiko Yanase}, \bibinfo{person}{Takeru Ohta}, {and} \bibinfo{person}{Masanori Koyama}.} \bibinfo{year}{2019}\natexlab{}.
\newblock \showarticletitle{Optuna: A Next-generation Hyperparameter Optimization Framework}. In \bibinfo{booktitle}{\emph{Proceedings of the 25th ACM SIGKDD International Conference on Knowledge Discovery \& Data Mining}} (Anchorage, AK, USA) \emph{(\bibinfo{series}{KDD '19})}. \bibinfo{publisher}{Association for Computing Machinery}, \bibinfo{address}{New York, NY, USA}, \bibinfo{pages}{2623–2631}.
\newblock
\showISBNx{9781450362016}


\bibitem[Bruch et~al\mbox{.}(2020)]%
        {10.1145/3336191.3371844}
\bibfield{author}{\bibinfo{person}{Sebastian Bruch}, \bibinfo{person}{Shuguang Han}, \bibinfo{person}{Michael Bendersky}, {and} \bibinfo{person}{Marc Najork}.} \bibinfo{year}{2020}\natexlab{}.
\newblock \showarticletitle{A Stochastic Treatment of Learning to Rank Scoring Functions}. In \bibinfo{booktitle}{\emph{Proceedings of the 13th International Conference on Web Search and Data Mining}} (Houston, TX, USA) \emph{(\bibinfo{series}{WSDM '20})}. \bibinfo{publisher}{Association for Computing Machinery}, \bibinfo{address}{New York, NY, USA}, \bibinfo{pages}{61–69}.
\newblock
\showISBNx{9781450368223}


\bibitem[Buchholz et~al\mbox{.}(2022)]%
        {Buchholz2022}
\bibfield{author}{\bibinfo{person}{Alexander Buchholz}, \bibinfo{person}{Jan~Malte Lichtenberg}, \bibinfo{person}{Giuseppe~Di Benedetto}, \bibinfo{person}{Yannik Stein}, \bibinfo{person}{Vito Bellini}, {and} \bibinfo{person}{Matteo Ruffini}.} \bibinfo{year}{2022}\natexlab{}.
\newblock \showarticletitle{Low-variance estimation in the Plackett-Luce model via quasi-Monte Carlo sampling}. In \bibinfo{booktitle}{\emph{SIGIR 2022 Workshop on Reaching Efficiency in Neural Information Retrieval}}.
\newblock


\bibitem[Burges(2010)]%
        {burges2010from}
\bibfield{author}{\bibinfo{person}{Chris~J.C. Burges}.} \bibinfo{year}{2010}\natexlab{}.
\newblock \bibinfo{booktitle}{\emph{From RankNet to LambdaRank to LambdaMART: An Overview}}.
\newblock \bibinfo{type}{{T}echnical {R}eport} MSR-TR-2010-82.
\newblock


\bibitem[Cao et~al\mbox{.}(2007)]%
        {cao2007learning}
\bibfield{author}{\bibinfo{person}{Zhe Cao}, \bibinfo{person}{Tao Qin}, \bibinfo{person}{Tie-Yan Liu}, \bibinfo{person}{Ming-Feng Tsai}, {and} \bibinfo{person}{Hang Li}.} \bibinfo{year}{2007}\natexlab{}.
\newblock \showarticletitle{Learning to rank: from pairwise approach to listwise approach}. In \bibinfo{booktitle}{\emph{Proceedings of the 24th International Conference on Machine Learning}} (Corvalis, Oregon, USA) \emph{(\bibinfo{series}{ICML '07})}. \bibinfo{publisher}{Association for Computing Machinery}, \bibinfo{address}{New York, NY, USA}, \bibinfo{pages}{129–136}.
\newblock
\showISBNx{9781595937933}


\bibitem[Chakrabarti et~al\mbox{.}(2008)]%
        {10.1145/1401890.1401906}
\bibfield{author}{\bibinfo{person}{Soumen Chakrabarti}, \bibinfo{person}{Rajiv Khanna}, \bibinfo{person}{Uma Sawant}, {and} \bibinfo{person}{Chiru Bhattacharyya}.} \bibinfo{year}{2008}\natexlab{}.
\newblock \showarticletitle{Structured Learning for Non-Smooth Ranking Losses}. In \bibinfo{booktitle}{\emph{Proceedings of the 14th ACM SIGKDD International Conference on Knowledge Discovery and Data Mining}} (Las Vegas, Nevada, USA) \emph{(\bibinfo{series}{KDD '08})}. \bibinfo{publisher}{Association for Computing Machinery}, \bibinfo{address}{New York, NY, USA}, \bibinfo{pages}{88–96}.
\newblock
\showISBNx{9781605581934}


\bibitem[Chapelle and Chang(2011)]%
        {pmlr-v14-chapelle11a}
\bibfield{author}{\bibinfo{person}{Olivier Chapelle} {and} \bibinfo{person}{Yi Chang}.} \bibinfo{year}{2011}\natexlab{}.
\newblock \showarticletitle{Yahoo! Learning to Rank Challenge Overview}. In \bibinfo{booktitle}{\emph{Proceedings of the Learning to Rank Challenge}} \emph{(\bibinfo{series}{Proceedings of Machine Learning Research}, Vol.~\bibinfo{volume}{14})}, \bibfield{editor}{\bibinfo{person}{Olivier Chapelle}, \bibinfo{person}{Yi~Chang}, {and} \bibinfo{person}{Tie-Yan Liu}} (Eds.). \bibinfo{publisher}{PMLR}, \bibinfo{address}{Haifa, Israel}, \bibinfo{pages}{1--24}.
\newblock


\bibitem[Chen and Guestrin(2016)]%
        {10.1145/2939672.2939785}
\bibfield{author}{\bibinfo{person}{Tianqi Chen} {and} \bibinfo{person}{Carlos Guestrin}.} \bibinfo{year}{2016}\natexlab{}.
\newblock \showarticletitle{XGBoost: A Scalable Tree Boosting System}. In \bibinfo{booktitle}{\emph{Proceedings of the 22nd ACM SIGKDD International Conference on Knowledge Discovery and Data Mining}} (San Francisco, California, USA) \emph{(\bibinfo{series}{KDD '16})}. \bibinfo{publisher}{Association for Computing Machinery}, \bibinfo{address}{New York, NY, USA}, \bibinfo{pages}{785–794}.
\newblock
\showISBNx{9781450342322}


\bibitem[Dato et~al\mbox{.}(2016)]%
        {10.1145/2987380}
\bibfield{author}{\bibinfo{person}{Domenico Dato}, \bibinfo{person}{Claudio Lucchese}, \bibinfo{person}{Franco~Maria Nardini}, \bibinfo{person}{Salvatore Orlando}, \bibinfo{person}{Raffaele Perego}, \bibinfo{person}{Nicola Tonellotto}, {and} \bibinfo{person}{Rossano Venturini}.} \bibinfo{year}{2016}\natexlab{}.
\newblock \showarticletitle{Fast Ranking with Additive Ensembles of Oblivious and Non-Oblivious Regression Trees}.
\newblock \bibinfo{journal}{\emph{ACM Trans. Inf. Syst.}} \bibinfo{volume}{35}, \bibinfo{number}{2}, Article \bibinfo{articleno}{15} (\bibinfo{date}{dec} \bibinfo{year}{2016}), \bibinfo{numpages}{31}~pages.
\newblock
\showISSN{1046-8188}


\bibitem[Dato et~al\mbox{.}(2022)]%
        {dato:sigir2022-istella}
\bibfield{author}{\bibinfo{person}{Domenico Dato}, \bibinfo{person}{Sean MacAvaney}, \bibinfo{person}{Franco~Maria Nardini}, \bibinfo{person}{Raffaele Perego}, {and} \bibinfo{person}{Nicola Tonellotto}.} \bibinfo{year}{2022}\natexlab{}.
\newblock \showarticletitle{The Istella22 Dataset: Bridging Traditional and Neural Learning to Rank Evaluation}. In \bibinfo{booktitle}{\emph{Proceedings of the 45th International ACM SIGIR Conference on Research and Development in Information Retrieval}} \emph{(\bibinfo{series}{SIGIR '22})}. \bibinfo{publisher}{Association for Computing Machinery}, \bibinfo{address}{New York, NY, USA}, \bibinfo{pages}{3099–3107}.
\newblock
\showISBNx{9781450387323}


\bibitem[Diaz et~al\mbox{.}(2020)]%
        {10.1145/3340531.3411962}
\bibfield{author}{\bibinfo{person}{Fernando Diaz}, \bibinfo{person}{Bhaskar Mitra}, \bibinfo{person}{Michael~D. Ekstrand}, \bibinfo{person}{Asia~J. Biega}, {and} \bibinfo{person}{Ben Carterette}.} \bibinfo{year}{2020}\natexlab{}.
\newblock \showarticletitle{Evaluating Stochastic Rankings with Expected Exposure}. In \bibinfo{booktitle}{\emph{Proceedings of the 29th ACM International Conference on Information \& Knowledge Management}} (Virtual Event, Ireland) \emph{(\bibinfo{series}{CIKM '20})}. \bibinfo{publisher}{Association for Computing Machinery}, \bibinfo{address}{New York, NY, USA}, \bibinfo{pages}{275–284}.
\newblock
\showISBNx{9781450368599}


\bibitem[Gumbel(1954)]%
        {gumbel1948statistical}
\bibfield{author}{\bibinfo{person}{Emil~Julius Gumbel}.} \bibinfo{year}{1954}\natexlab{}.
\newblock \showarticletitle{Statistical Theory of Extreme Values and Some Practical Applications}.
\newblock \bibinfo{journal}{\emph{National Bureau of Standards Applied Mathematics Series. 33}} (\bibinfo{year}{1954}).
\newblock


\bibitem[Harman(2011)]%
        {harman2011information}
\bibfield{author}{\bibinfo{person}{Donna Harman}.} \bibinfo{year}{2011}\natexlab{}.
\newblock \bibinfo{booktitle}{\emph{Information Retrieval Evaluation}}.
\newblock \bibinfo{publisher}{Morgan \& Claypool Publishers}.
\newblock


\bibitem[Jagerman et~al\mbox{.}(2022)]%
        {jagerman2022optimizing}
\bibfield{author}{\bibinfo{person}{Rolf Jagerman}, \bibinfo{person}{Zhen Qin}, \bibinfo{person}{Xuanhui Wang}, \bibinfo{person}{Michael Bendersky}, {and} \bibinfo{person}{Marc Najork}.} \bibinfo{year}{2022}\natexlab{}.
\newblock \showarticletitle{On Optimizing Top-K Metrics for Neural Ranking Models}. In \bibinfo{booktitle}{\emph{Proceedings of the 45th International ACM SIGIR Conference on Research and Development in Information Retrieval}} \emph{(\bibinfo{series}{SIGIR '22})}. \bibinfo{publisher}{Association for Computing Machinery}, \bibinfo{address}{New York, NY, USA}, \bibinfo{pages}{2303–2307}.
\newblock
\showISBNx{9781450387323}


\bibitem[J\"{a}rvelin and Kek\"{a}l\"{a}inen(2002)]%
        {jarvelin2002cumulated}
\bibfield{author}{\bibinfo{person}{Kalervo J\"{a}rvelin} {and} \bibinfo{person}{Jaana Kek\"{a}l\"{a}inen}.} \bibinfo{year}{2002}\natexlab{}.
\newblock \showarticletitle{Cumulated gain-based evaluation of IR techniques}.
\newblock \bibinfo{journal}{\emph{ACM Trans. Inf. Syst.}} \bibinfo{volume}{20}, \bibinfo{number}{4} (\bibinfo{date}{oct} \bibinfo{year}{2002}), \bibinfo{pages}{422–446}.
\newblock
\showISSN{1046-8188}


\bibitem[Liu(2009)]%
        {INR-016}
\bibfield{author}{\bibinfo{person}{Tie-Yan Liu}.} \bibinfo{year}{2009}\natexlab{}.
\newblock \showarticletitle{Learning to Rank for Information Retrieval}.
\newblock \bibinfo{journal}{\emph{Foundations and Trends® in Information Retrieval}} \bibinfo{volume}{3}, \bibinfo{number}{3} (\bibinfo{year}{2009}), \bibinfo{pages}{225--331}.
\newblock


\bibitem[Luce(1959)]%
        {luce_individual_1959}
\bibfield{author}{\bibinfo{person}{R.~Duncan Luce}.} \bibinfo{year}{1959}\natexlab{}.
\newblock \bibinfo{booktitle}{\emph{Individual Choice Behavior: A Theoretical Analysis}}.
\newblock \bibinfo{publisher}{Wiley}.
\newblock


\bibitem[Oosterhuis(2021)]%
        {10.1145/3404835.3462830}
\bibfield{author}{\bibinfo{person}{Harrie Oosterhuis}.} \bibinfo{year}{2021}\natexlab{}.
\newblock \showarticletitle{Computationally Efficient Optimization of Plackett-Luce Ranking Models for Relevance and Fairness}. In \bibinfo{booktitle}{\emph{Proceedings of the 44th International ACM SIGIR Conference on Research and Development in Information Retrieval}} (Virtual Event, Canada) \emph{(\bibinfo{series}{SIGIR '21})}. \bibinfo{publisher}{Association for Computing Machinery}, \bibinfo{address}{New York, NY, USA}, \bibinfo{pages}{1023–1032}.
\newblock
\showISBNx{9781450380379}


\bibitem[Oosterhuis(2022)]%
        {10.1145/3477495.3531842}
\bibfield{author}{\bibinfo{person}{Harrie Oosterhuis}.} \bibinfo{year}{2022}\natexlab{}.
\newblock \showarticletitle{Learning-to-Rank at the Speed of Sampling: Plackett-Luce Gradient Estimation with Minimal Computational Complexity}. In \bibinfo{booktitle}{\emph{Proceedings of the 45th International ACM SIGIR Conference on Research and Development in Information Retrieval}} (Madrid, Spain) \emph{(\bibinfo{series}{SIGIR '22})}. \bibinfo{publisher}{Association for Computing Machinery}, \bibinfo{address}{New York, NY, USA}, \bibinfo{pages}{2266–2271}.
\newblock
\showISBNx{9781450387323}


\bibitem[Paszke et~al\mbox{.}(2019)]%
        {NEURIPS2019_bdbca288}
\bibfield{author}{\bibinfo{person}{Adam Paszke}, \bibinfo{person}{Sam Gross}, \bibinfo{person}{Francisco Massa}, \bibinfo{person}{Adam Lerer}, \bibinfo{person}{James Bradbury}, \bibinfo{person}{Gregory Chanan}, \bibinfo{person}{Trevor Killeen}, \bibinfo{person}{Zeming Lin}, \bibinfo{person}{Natalia Gimelshein}, \bibinfo{person}{Luca Antiga}, \bibinfo{person}{Alban Desmaison}, \bibinfo{person}{Andreas Kopf}, \bibinfo{person}{Edward Yang}, \bibinfo{person}{Zachary DeVito}, \bibinfo{person}{Martin Raison}, \bibinfo{person}{Alykhan Tejani}, \bibinfo{person}{Sasank Chilamkurthy}, \bibinfo{person}{Benoit Steiner}, \bibinfo{person}{Lu Fang}, \bibinfo{person}{Junjie Bai}, {and} \bibinfo{person}{Soumith Chintala}.} \bibinfo{year}{2019}\natexlab{}.
\newblock \showarticletitle{PyTorch: An Imperative Style, High-Performance Deep Learning Library}. In \bibinfo{booktitle}{\emph{Advances in Neural Information Processing Systems}}, \bibfield{editor}{\bibinfo{person}{H.~Wallach}, \bibinfo{person}{H.~Larochelle}, \bibinfo{person}{A.~Beygelzimer}, \bibinfo{person}{F.~d\textquotesingle Alch\'{e}-Buc}, \bibinfo{person}{E.~Fox}, {and} \bibinfo{person}{R.~Garnett}} (Eds.), Vol.~\bibinfo{volume}{32}. \bibinfo{publisher}{Curran Associates, Inc.}, \bibinfo{address}{Red Hook, NY, USA}, \bibinfo{pages}{12}.
\newblock


\bibitem[Plackett(1975)]%
        {af5079a1-8ca5-3727-a405-0a82390327b7}
\bibfield{author}{\bibinfo{person}{R.~L. Plackett}.} \bibinfo{year}{1975}\natexlab{}.
\newblock \showarticletitle{The Analysis of Permutations}.
\newblock \bibinfo{journal}{\emph{Journal of the Royal Statistical Society. Series C (Applied Statistics)}} \bibinfo{volume}{24}, \bibinfo{number}{2} (\bibinfo{year}{1975}), \bibinfo{pages}{193--202}.
\newblock
\showISSN{00359254, 14679876}


\bibitem[Qin and Liu(2013)]%
        {DBLP:journals/corr/QinL13}
\bibfield{author}{\bibinfo{person}{Tao Qin} {and} \bibinfo{person}{Tie{-}Yan Liu}.} \bibinfo{year}{2013}\natexlab{}.
\newblock \showarticletitle{Introducing {LETOR} 4.0 Datasets}.
\newblock \bibinfo{journal}{\emph{CoRR}}  \bibinfo{volume}{abs/1306.2597} (\bibinfo{year}{2013}).
\newblock


\bibitem[Qin et~al\mbox{.}(2010)]%
        {qin2010general}
\bibfield{author}{\bibinfo{person}{Tao Qin}, \bibinfo{person}{Tie-Yan Liu}, {and} \bibinfo{person}{Hang Li}.} \bibinfo{year}{2010}\natexlab{}.
\newblock \showarticletitle{A General Approximation Framework for Direct Optimization of Information Retrieval Measures}.
\newblock \bibinfo{journal}{\emph{Information retrieval}}  \bibinfo{volume}{13} (\bibinfo{year}{2010}), \bibinfo{pages}{375--397}.
\newblock


\bibitem[Qin et~al\mbox{.}(2021)]%
        {50030}
\bibfield{author}{\bibinfo{person}{Zhen Qin}, \bibinfo{person}{Le Yan}, \bibinfo{person}{Honglei Zhuang}, \bibinfo{person}{Yi Tay}, \bibinfo{person}{Rama~Kumar Pasumarthi}, \bibinfo{person}{Xuanhui Wang}, \bibinfo{person}{Mike Bendersky}, {and} \bibinfo{person}{Marc Najork}.} \bibinfo{year}{2021}\natexlab{}.
\newblock \showarticletitle{Are Neural Rankers still Outperformed by Gradient Boosted Decision Trees?}. In \bibinfo{booktitle}{\emph{International Conference on Learning Representations (ICLR)}}.
\newblock


\bibitem[Singh and Joachims(2019)]%
        {NEURIPS2019_9e82757e}
\bibfield{author}{\bibinfo{person}{Ashudeep Singh} {and} \bibinfo{person}{Thorsten Joachims}.} \bibinfo{year}{2019}\natexlab{}.
\newblock \bibinfo{booktitle}{\emph{Policy learning for fairness in ranking}}.
\newblock \bibinfo{publisher}{Curran Associates Inc.}, \bibinfo{address}{Red Hook, NY, USA}.
\newblock


\bibitem[Taylor et~al\mbox{.}(2008)]%
        {10.1145/1341531.1341544}
\bibfield{author}{\bibinfo{person}{Michael Taylor}, \bibinfo{person}{John Guiver}, \bibinfo{person}{Stephen Robertson}, {and} \bibinfo{person}{Tom Minka}.} \bibinfo{year}{2008}\natexlab{}.
\newblock \showarticletitle{SoftRank: optimizing non-smooth rank metrics}. In \bibinfo{booktitle}{\emph{Proceedings of the 2008 International Conference on Web Search and Data Mining}} (Palo Alto, California, USA) \emph{(\bibinfo{series}{WSDM '08})}. \bibinfo{publisher}{Association for Computing Machinery}, \bibinfo{address}{New York, NY, USA}, \bibinfo{pages}{77–86}.
\newblock
\showISBNx{9781595939272}


\bibitem[Ustimenko and Prokhorenkova(2020)]%
        {ustimenko2020stochasticrank}
\bibfield{author}{\bibinfo{person}{Aleksei Ustimenko} {and} \bibinfo{person}{Liudmila Prokhorenkova}.} \bibinfo{year}{2020}\natexlab{}.
\newblock \showarticletitle{{S}tochastic{R}ank: Global Optimization of Scale-Free Discrete Functions}. In \bibinfo{booktitle}{\emph{Proceedings of the 37th International Conference on Machine Learning}} \emph{(\bibinfo{series}{Proceedings of Machine Learning Research}, Vol.~\bibinfo{volume}{119})}. \bibinfo{publisher}{PMLR}, \bibinfo{pages}{9669--9679}.
\newblock


\bibitem[Xia et~al\mbox{.}(2008)]%
        {xia2008listwise}
\bibfield{author}{\bibinfo{person}{Fen Xia}, \bibinfo{person}{Tie-Yan Liu}, \bibinfo{person}{Jue Wang}, \bibinfo{person}{Wensheng Zhang}, {and} \bibinfo{person}{Hang Li}.} \bibinfo{year}{2008}\natexlab{}.
\newblock \showarticletitle{Listwise approach to learning to rank: theory and algorithm}. In \bibinfo{booktitle}{\emph{Proceedings of the 25th International Conference on Machine Learning}} (Helsinki, Finland) \emph{(\bibinfo{series}{ICML '08})}. \bibinfo{publisher}{Association for Computing Machinery}, \bibinfo{address}{New York, NY, USA}, \bibinfo{pages}{1192–1199}.
\newblock
\showISBNx{9781605582054}


\bibitem[Xu et~al\mbox{.}(2008)]%
        {10.1145/1390334.1390355}
\bibfield{author}{\bibinfo{person}{Jun Xu}, \bibinfo{person}{Tie-Yan Liu}, \bibinfo{person}{Min Lu}, \bibinfo{person}{Hang Li}, {and} \bibinfo{person}{Wei-Ying Ma}.} \bibinfo{year}{2008}\natexlab{}.
\newblock \showarticletitle{Directly Optimizing Evaluation Measures in Learning to Rank}. In \bibinfo{booktitle}{\emph{Proceedings of the 31st Annual International ACM SIGIR Conference on Research and Development in Information Retrieval}} (Singapore, Singapore) \emph{(\bibinfo{series}{SIGIR '08})}. \bibinfo{publisher}{Association for Computing Machinery}, \bibinfo{address}{New York, NY, USA}, \bibinfo{pages}{107–114}.
\newblock
\showISBNx{9781605581644}


\bibitem[Yue et~al\mbox{.}(2007)]%
        {10.1145/1277741.1277790}
\bibfield{author}{\bibinfo{person}{Yisong Yue}, \bibinfo{person}{Thomas Finley}, \bibinfo{person}{Filip Radlinski}, {and} \bibinfo{person}{Thorsten Joachims}.} \bibinfo{year}{2007}\natexlab{}.
\newblock \showarticletitle{A Support Vector Method for Optimizing Average Precision}. In \bibinfo{booktitle}{\emph{Proceedings of the 30th Annual International ACM SIGIR Conference on Research and Development in Information Retrieval}} (Amsterdam, The Netherlands) \emph{(\bibinfo{series}{SIGIR '07})}. \bibinfo{publisher}{Association for Computing Machinery}, \bibinfo{address}{New York, NY, USA}, \bibinfo{pages}{271–278}.
\newblock
\showISBNx{9781595935977}


\end{thebibliography}

\end{document}